\documentclass{article}

\usepackage{microtype}
\usepackage{graphicx}
\usepackage{subfigure}
\usepackage{booktabs} 
\usepackage[ruled,lined]{algorithm2e}
\usepackage{algorithm} 
\usepackage{algorithmic}
\usepackage{multirow}

\usepackage{hyperref}

\usepackage[accepted]{mlsys2023}
\usepackage{multicol,lipsum}
\usepackage{makecell}
\graphicspath{ {./figures/} }

\mlsystitlerunning{An Efficient Sparse Inference Software Accelerator for Transformer-based Language Models on CPUs}

\begin{document}

\twocolumn[
\mlsystitle{An Efficient Sparse Inference Software Accelerator for Transformer-based Language Models on CPUs}




\begin{mlsysauthorlist}
\mlsysauthor{Haihao Shen}{intel}
\mlsysauthor{Hengyu Meng}{intel}
\mlsysauthor{Bo Dong}{intel}
\mlsysauthor{Zhe Wang}{intel}
\mlsysauthor{Ofir Zafrir}{labs}
\mlsysauthor{Yi Ding}{intel}
\mlsysauthor{Yu Luo}{intel}
\mlsysauthor{Hanwen Chang}{intel}
\mlsysauthor{Qun Gao}{intel}
\mlsysauthor{Ziheng Wang}{stanford}
\mlsysauthor{Guy Boudoukh}{labs}
\mlsysauthor{Moshe Wasserblat}{labs}
\end{mlsysauthorlist}

\mlsysaffiliation{intel}{Intel}
\mlsysaffiliation{labs}{Intel Labs}
\mlsysaffiliation{stanford}{Stanford University}
\vskip 0.1in
\centering{\textbf{\footnote{1}Intel~~~~~~\footnote{2}Intel Labs~~~~~~ \footnote{3}Stanford University}}



\vskip 0.2in

\begin{abstract}
In recent years, Transformer-based language models have become the standard approach for natural language processing tasks. However, stringent throughput and latency requirements in industrial applications are limiting their adoption. To mitigate the gap, model compression techniques such as structured pruning are being used to improve inference efficiency. However, most existing neural network inference runtimes lack adequate support for structured sparsity. In this paper, we propose an efficient sparse deep learning inference software stack for Transformer-based language models where the weights are pruned with constant block size. Our sparse software accelerator leverages Intel\textsuperscript{\tiny\textregistered} Deep Learning Boost to maximize the performance of sparse matrix - dense matrix multiplication (commonly abbreviated as SpMM) on CPUs. Our SpMM kernel outperforms the existing sparse libraries (oneMKL, TVM, and LIBXSMM) by an order of magnitude on a wide range of GEMM shapes under 5 representative sparsity ratios (70\%, 75\%, 80\%, 85\%, 90\%). Moreover, our SpMM kernel shows up to 5x speedup over dense GEMM kernel of oneDNN, a well-optimized dense library widely used in industry. We apply our sparse accelerator on widely-used Transformer-based language models including Bert-Mini, DistilBERT, Bert-Base, and BERT-Large. Our sparse inference software shows up to 1.5x speedup over Neural Magic's Deepsparse under same configurations on Xeon on Amazon Web Services under proxy production latency constraints. We also compare our solution with two framework-based inference solutions, ONNX Runtime and PyTorch, and demonstrate up to 37x speedup over ONNX Runtime and 345x over PyTorch on Xeon under the latency constraints. All the source code is publicly available on Github~\footnote{https://github.com/intel/intel-extension-for-transformers}.
\end{abstract}
]




\section{Introduction}
\label{sec:intro}
Large Transformer-based Language Models (LMs) are evolving rapidly from millions of parameters, e.g., BERT-Large~\cite{devlin2018bert}, to billions of parameters, e.g., Turing-Megatron~\cite{smith2022using}, and GPT3~\cite{Brown2020LanguageMA}. Transformer-based LMs are currently used for solving almost all natural language processing (NLP) tasks, and those large models have demonstrated promising state-of-the-art (SoTA) accuracy on a wide range of NLP tasks. However, it's still quite challenging to deploy these models in production due to the demands of large computation resources and strict latency constraints in real applications.

To enable the deployment of Transformer models at scale, model compression and optimization are usually required to make model inference more efficient. Pruning~\cite{sanh2020movement} and quantization~\cite{zafrir2019q8bert} are two well-known approaches that have been widely used to compress Transformer models. There are two kinds of pruning methods: unstructured pruning~\cite{han2015learning}~\cite{gordon2020compressing}~\cite{wang2020sparsert} and structured pruning~\cite{pool2021accelerating}, where unstructured pruning does not require any special sparsity pattern while structured one requires applying the same sparsity pattern in the weights. In general, it is difficult to demonstrate the performance gains for an unstructured sparse model without high sparsity ratio. Even for a structured sparse model, speedup might be difficult without special hardware support (e.g., NVIDIA Ampere architecture and beyond). On the other hand, there are some recent works~\cite{yao2022zeroquant}~\cite{kim2021bert} that have demonstrated the performance of 8-bit integer (INT8) quantization as 8-bit fused multiply-add (FMA) or matrix multiplication instruction has been supported in majority modern hardwares. However, there is a lack of work that demonstrate the efficient inference on modern CPUs combining quantization and pruning.

In this paper, we propose an efficient sparse deep learning inference software stack for Transformer-based LMs that supports efficient structured sparsity with quantization. We define a structured sparsity pattern based on constant block size 4x1. We generate the sparse model based on the sparsity pattern and further quantize the sparse model to make the inference more efficient while maintaining the acceptable accuracy. To demonstrate the performance of a quantized sparse model, we implement SpMM kernel based on Intel\textsuperscript{\tiny\textregistered} Deep Learning Boost, as well as a sparse attention kernel. We measure the performance of SpMM kernel on a wide range of typical shapes (totally 90) under 5 representative sparsity ratios (70\%, 75\%, 80\%, 85\%, 90\%), and compare with (structured/unstructured) sparse GEMM kernel on popular libraries (oneMKL, TVM, and LIBXSMM) and dense GEMM kernel on oneDNN. Our SpMM kernel outperforms the existing sparse libraries on the performance by an order of magnitude. In particular, we compare structured SpMM with same block size (namely same number of block elements) e.g., 4x1 and 2x2), it shows the speedup up to 17x over oneMKL and up to 41x over TVM on single thread, and up to 20x over oneMKL and up to 62x over TVM on four threads. We also compare with dense GEMM kernel of oneDNN and demonstrate up to 5x performance. Moreover, our SpMM kernel shows almost linear scaling from single thread to four threads among all the configurations of different shapes and sparsity ratios. We apply the sparse accelerator on popular widely-used Transformer-based language models including Bert-Mini, DistilBERT, Bert-Base, and BERT-Large. We also compare the workload performance with Neural Magic's Deepsparse (a leading sparsity-aware inference engine)\footnote{https://github.com/neuralmagic/deepsparse} and demonstrate up to 1.5x speedup on same CPU instance on Xeon and up to 4.9x on different CPU instances (Xeon for Sparse Accelerator vs. Eypc for Neural Magic) respectively on Amazon Web Services (AWS) under the proxy production latency constraints. We also compare our solution with ONNX Runtime and PyTorch (framework-based inference solution) and demonstrate the speedup up to 37x over ONNX Runtime and 345x over PyTorch on same Xeon instance, and 72x over ONNX Runtime and 309x over PyTorch on Xeon vs. Eypc instances under the latency constraint. In summary, our main contributions are:
\begin{itemize}
  \item Define a structured sparsity pattern with block size 4x1 and generate 9 sparse Transformer models on various downstream tasks with 80\% - 90\% sparsity ratio and prove the accuracy within 1\% loss from baseline.
  \item Develop efficient SpMM and sparse attention techniques based on Intel\textsuperscript{\tiny\textregistered} Deep Learning Boost for Transformer-based LMs.
  \item Outperform existing libraries (oneMKL, TVM, and LIBXSMM) on SpMM kernel performance by an order of magnitude: up to 17x over oneMKL and 41x over TVM on single thread; up to 20x over oneMKL and 62x over TVM on multi-threads. Show up to 5x performance over dense GEMM kernel of oneDNN.
  \item Demonstrate good end-to-end speedup: up to 1.5x to 4.9x over Neural Magic from same to different instances (Xeon for sparse accelerator vs. Eypc for Neural Magic); up to 37x - 72x over ONNX Runtime and 309x - 345x over PyTorch from Xeon to Eypc instance under the latency constraint.
\end{itemize}

\section{Related Work}
\label{sec:related_work}

\subsection{Model Compression}
Transformer-based LMs have demonstrated SoTA accuracy on a variety range of NLP tasks while the model size is growing rapidly. However, those models are hard to deploy for production due to the limited computation resources and strict latency constraints. There has been a growing interest in the compression of Transformer-based LMs to improve the inference efficiency.

Pruning has been proven to be an effective way of reducing model size while maintaining the similar model quality~\cite{lecun1989optimal}~\cite{sanh2020movement}~\cite{wang2021sparsednn}. Structured pruning is gaining popularity to prune the weights with a pre-defined sparsity pattern such as block-wise pruning~\cite{lagunas2021block} and fine-grained 2:4~\cite{pool2021channel} or N:M structured sparsity~\cite{zhou2021learning}. Recent works~\cite{zafrir2021prune, kurtic2022optimal} proposed pruning Transformer models at pre-training to create sparse pre-trained LMs and fine-tuning on downstream tasks.

Quantization is another widely-used model compression technique that can improve the inference latency~\cite{jacob2018quantization}\cite{zafrir2019q8bert}~\cite{bhandare2019efficient}. There are two typical quantization approaches: post-training quantization (PTQ) and quantization-aware training (QAT), where PTQ requires an offline calibration process on representative samples to collect the tensor statistics and generate the scale and zero point used for quantization, and QAT requires an additional fine-tuning phase simulating the quantization inference during training. 

Knowledge distillation is a popular compression technique~\cite{Hinton2015DistillingTK}~\cite{sanh2019distilbert}~\cite{tang2019distilling}. It has been used to produce a much smaller BERT model~\cite{jiao2019tinybert}~\cite{sun2020mobilebert} while achieving high accuracy. Typically, distillation can be incorporated into pruning and quantization as a combined orchestrated model compression technique~\cite{zafrir2021prune}~\cite{yao2022zeroquant} which can produce a compressed model with the best trade-off among model size, performance, and accuracy.

\subsection{Sparse/Dense GEMM Libraries}
There are several existing sparse and dense GEMM libraries that support CPUs and/or GPUs.

oneAPI Math Kernel Library (oneMKL for short)\footnote{https://github.com/oneapi-src/oneMKL} has supported dense and sparse GEMM for a while. In particular for sparse GEMM, oneMKL supports multiple sparse matrix representations such as COO, CSR, BSR. However, sparse GEMM in oneMKL only supports 32-bit floating-point (FP32) data type and square block size such as 2x2.

LIBXSMM\footnote{https://github.com/libxsmm/libxsmm} is an open-source high performance library for small matrix multiplications. It supports both dense and unstructured sparse GEMM and demonstrates the impressive performance while it may require additional tuning to achieve the high performance. One of the constraints for LIBXSMM is that each dimension (M, K, N) of GEMM requires 32 dividable, which is limiting the usage for smaller shapes e.g., N = 16,

Apache TVM (TVM for short)\footnote{https://github.com/apache/tvm} is a widely-used compiler stack for deep learning systems which is designed to close the gap between productivity on deep learning frameworks and performance/efficiency on hardware backends. TVM supports two structured sparsity patterns (4x1 and 16x1), two sparse matrix representations (CSR and BSR), and two data types (FP32 and INT8).

oneAPI Deep Neural Network Library (oneDNN for short)~\footnote{https://github.com/oneapi-src/oneDNN} provides the mature support of dense GEMM kernels on CPU. It has been used in mainstream deep learning frameworks such as TensorFlow and PyTorch.

cuSparse\footnote{https://docs.nvidia.com/cuda/cusparse/index.html} is a sparse GEMM libary for CUDA, supporting unstructured and structured sparsity 2:4 recently introduced in NVidia Ampere architecture and above. hipSPARSE\footnote{https://github.com/ROCmSoftwarePlatform/hipSPARSE} is a sparse library created by AMD for their GPUs.

\subsection{Inference Solutions}
Modern deep learning frameworks such as TensorFlow, PyTorch, ONNX Runtime provide both training and inference to allow researchers and developers easily develop, build, and deploy AI models. In addition to these frameworks, OpenVINO\footnote{https://github.com/openvinotoolkit/openvino} is an open-source toolkit for optimizing and deploying AI inference, taking a model trained on frameworks as input and converting to its own intermediate representation (IR) as a pre-condition for deployment. However, these solutions do not support sparse models.

Besides the general frameworks or inference toolkits that support different hardware backends (e.g., CPU, GPU), there are some specialized inference solutions for dedicated hardwares. Neural Magic is a close-source sparsity-aware inference engine on CPU. It supports sparse GEMM kernel for both unstructured and structured sparsity and accelerates the performance on both Xeon and Eypc. To the best of our knowledge, this work is most relevant to ours which focuses on sparse model inference acceleration on CPUs. TensorRT\footnote{https://developer.nvidia.com/tensorrt} is a popular inference engine delivering the latest performance on NVidia hardwares, FasterTransformer\footnote{https://github.com/NVIDIA/FasterTransformer} is an accelerator for Transformer-based LMs by leveraging NVidia's 2:4 structured sparsity. 

Despite the popularity of GPUs, to the best of our knowledge, most industry inference is still done on CPUs, so the benchmark target in this paper is CPUs.

\section{Sparse Software Accelerator}
\label{sec:sparse_accelerator}
In this section, we present our sparse deep learning inference software accelerator for Transformer-based LMs, including structured sparsity pattern, sparse GEMM and Transformer attention kernels, and end to end optimizations.

\subsection{Sparsity Pattern}
\label{sec:sparse_pattern}
Choosing the right sparsity pattern is critical to achieving inference speedups while maintaining accuracy. There are two main factors that we need to consider: 1) structured vs. unstructured 2) sparsity pattern if structured. Note that we concentrate on weight sparsity only in this paper.

One of the main purpose in this work is to showcase efficient sparse inference on CPUs, and our target hardware is Intel\textsuperscript{\tiny\textregistered} Xeon\textsuperscript{\tiny\textregistered} Scalable Processors due to the support of advanced vector neural network instructions (VNNI) that can be used to maximize the performance of structured sparse kernels. We thus choose structured sparsity to make the best use of our target hardware.

\begin{figure}[htbp]
\centering
    \includegraphics[scale=0.22]{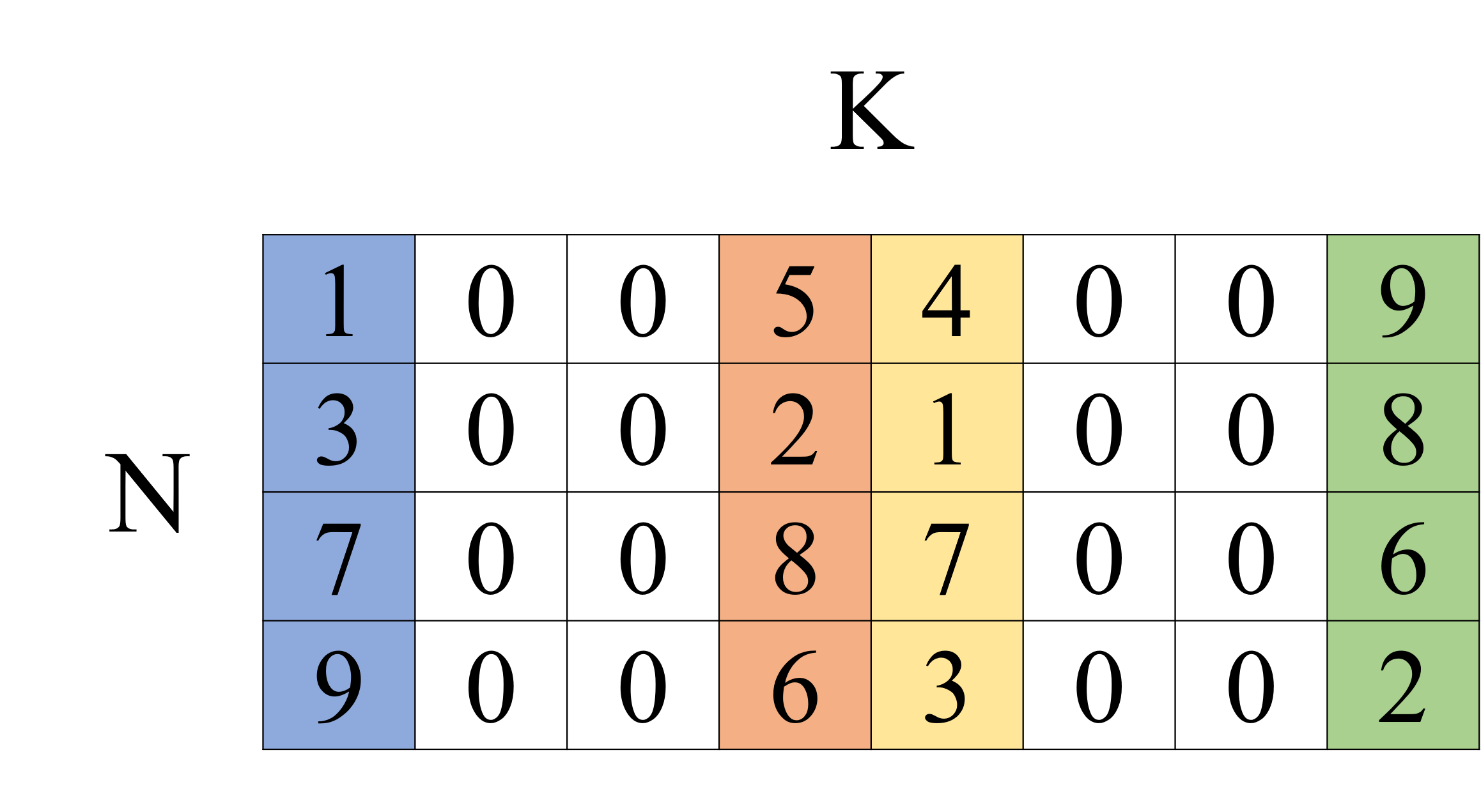}
    \caption{A sample sparse weight with structured sparsity pattern 4x1, where sparsity dimension is on N with highlighted colors}
    \label{fig:pattern}
\end{figure}

The next challenge is choosing a structured sparsity pattern. N:M such as 2:4 is out of our considerations, as there is lack of instruction support on our target hardware. We also exclude the option of a square block size (e.g., 2x2, 4x4) which leads to non-contiguous memory accesses. In this paper, we focus on 4x1 which offers the best trade-off among the time to train a sparse model, accuracy, and performance after extensive experiments. Figure~\ref{fig:pattern} shows a sample sparse weight with sparsity pattern 4x1.

\onecolumn
\begin{figure}[htbp]
    \centering
    \includegraphics[scale=0.6]{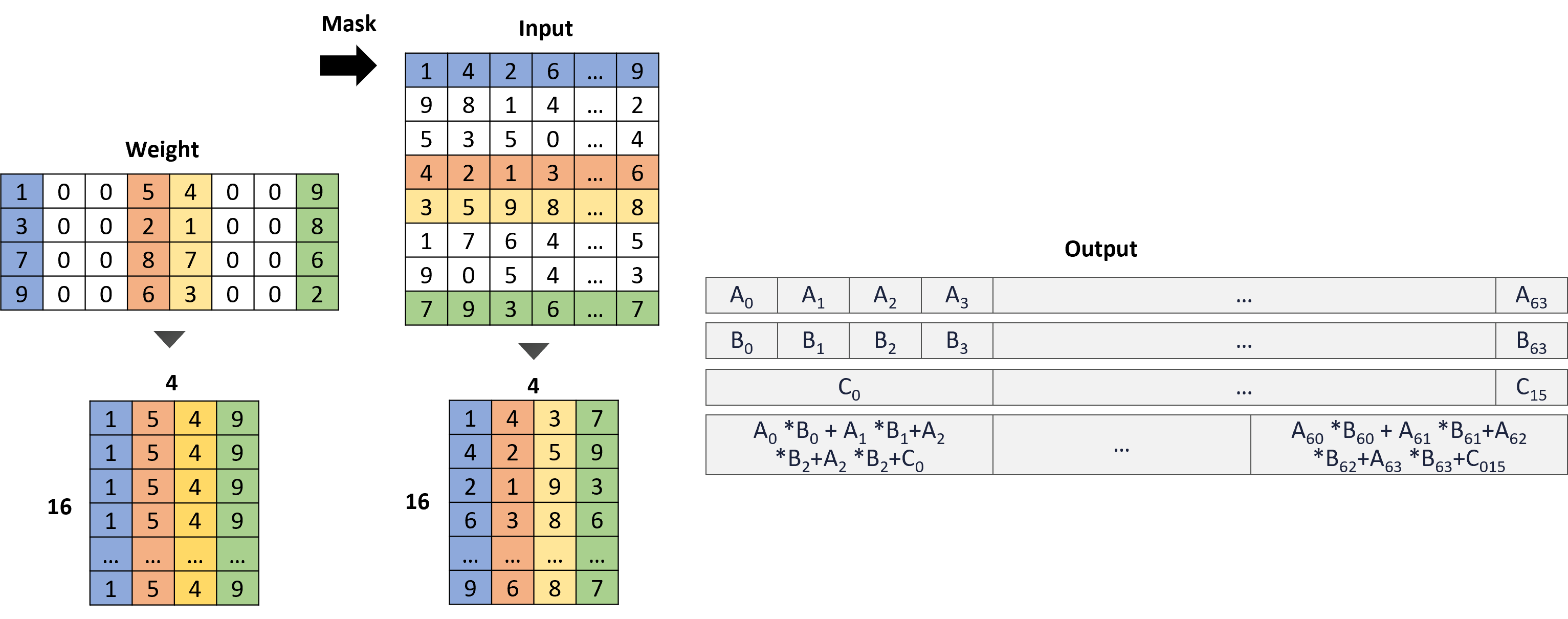}
    \caption{INT8 sparse GEMM kernel (sparse weight x dense input) implementation based on VNNI. Non-zero weight elements are broadcast to form a VNNI-format block (4x16), and input elements are re-structured per the mask of sparse weight to form another VNNI-format block. These two VNNI blocks are fed into VNNI to complete the sparse GEMM.}
    \label{fig:sparse}
\end{figure}

\begin{multicols}{2}

\subsection{Sparse GEMM Kernel}
\label{sec:sparse_gemm}
To demonstrate the performance of our defined sparsity pattern, we implement the sparse GEMM kernel by leveraging just-in-time (JIT) compilation to generate the machine code directly based on GEMM shapes, which gives the flexibility of bypassing the compiler to perform the loop unfolding more efficiently. Our JIT-based kernel implementation is especially useful for sparse GEMM kernels, since each kernel is specialized for a sparse weight tensor and sparse indices. We can just load the needed sparse indices before the kernel execution to save the instruction routing cost during the inference.

Given a sparse weight block NxK (sparsity pattern 4x1) and a dense input block, we first broadcast the non-zero weight block to form a VNNI-format block A. Based on the mask in the sparse weight, we re-structure the corresponding input as another VNNI-format block B on the fly based on AVX512 permutation and shuffling instructions. Then the kernel uses VNNI to produce the intermediate output given A and B, and add bias C as the final output. Algorithm~\ref{alg:sparse} describes the code snippet of INT8 sparse GEMM kernel with default optimization configurations. VNNI instructions are designed to multiply 16 groups of 4 adjacent pairs of unsigned 8-bit integers in one matrix with signed or unsigned 8-bit integers in the other matrix, produce 16 groups of 4 intermediate signed 16-bit results, add up these 4 results in the same group with 32-bit integer in destination matrix, and store the packed 32-bit results back in destination. This also explains why we use the constant block size 4 as our structured sparsity pattern, since 4 is the maximum tiling size to fully utilize the computation throughput in case no implicit register reusing, thereby improving the GEMM performance greatly. In particular, we apply tiling along N dimensions with \textit{n\_tile = 64} while corresponds to 4 times of VNNI width as the default configuration. Note that the tiling size is tunable offline to achieve even better performance given a GEMM shape with sparsity ratio.

\begin{algorithm}[H]
\caption{Code snippet of INT8 sparse GEMM kernel}
\label{alg:sparse}
// $M, N, K$ as three dimensions of GEMM

// $m\_block$ = 4, $n\_block$ = 64, $k\_block$ = 4

// $weight\_ptr$: weight tensor; $src\_ptr$: input tensor

    \For {$m = 0; m < M; m += m\_block$}{  
\For{$n = 0; n < N; n += n\_block$}{  
\For{$k = 0; k <= K; k += k\_block$}{
$vbroadcastss (\_m32i (weight\_ptr))$\;
$vbroadcastss (\_m32i (weight\_ptr))$\;
$vbroadcastss (\_m32i (weight\_ptr))$\;
$vbroadcastss (\_m32i (weight\_ptr))$\;

\For{$i = 0; i < 4; ++i$}{
$vmovdqu8 (\_m128i, src\_ptr)$\;
$vmovdqu8 (\_m128i, src\_ptr)$\;
$vbroadcasti32x4 (\_m512i, \_m128i)$\;
$vbroadcasti32x4 (\_m512i, \_m128i)$\;
$vpermt2d (\_m512i, \_m512i, \_m512i)$\;
$vpshufb (\_m512i, \_m512i, \_m512i)$\;
}
$vpdpbusd (\_m512i, \_m512i, \_m512i)$\;
$vpdpbusd (\_m512i, \_m512i, \_m512i)$\;
$vpdpbusd (\_m512i, \_m512i, \_m512i)$\;
$vpdpbusd (\_m512i, \_m512i, \_m512i)$\;

// downconvert and post-operator fusion\;
}   
}} 
\end{algorithm}

\end{multicols}

\onecolumn
\begin{figure}[htbp]
    \centering
        
    \subfigure[]{
        \begin{minipage}[t]{\linewidth}
            \centering
            \includegraphics[scale=0.22]{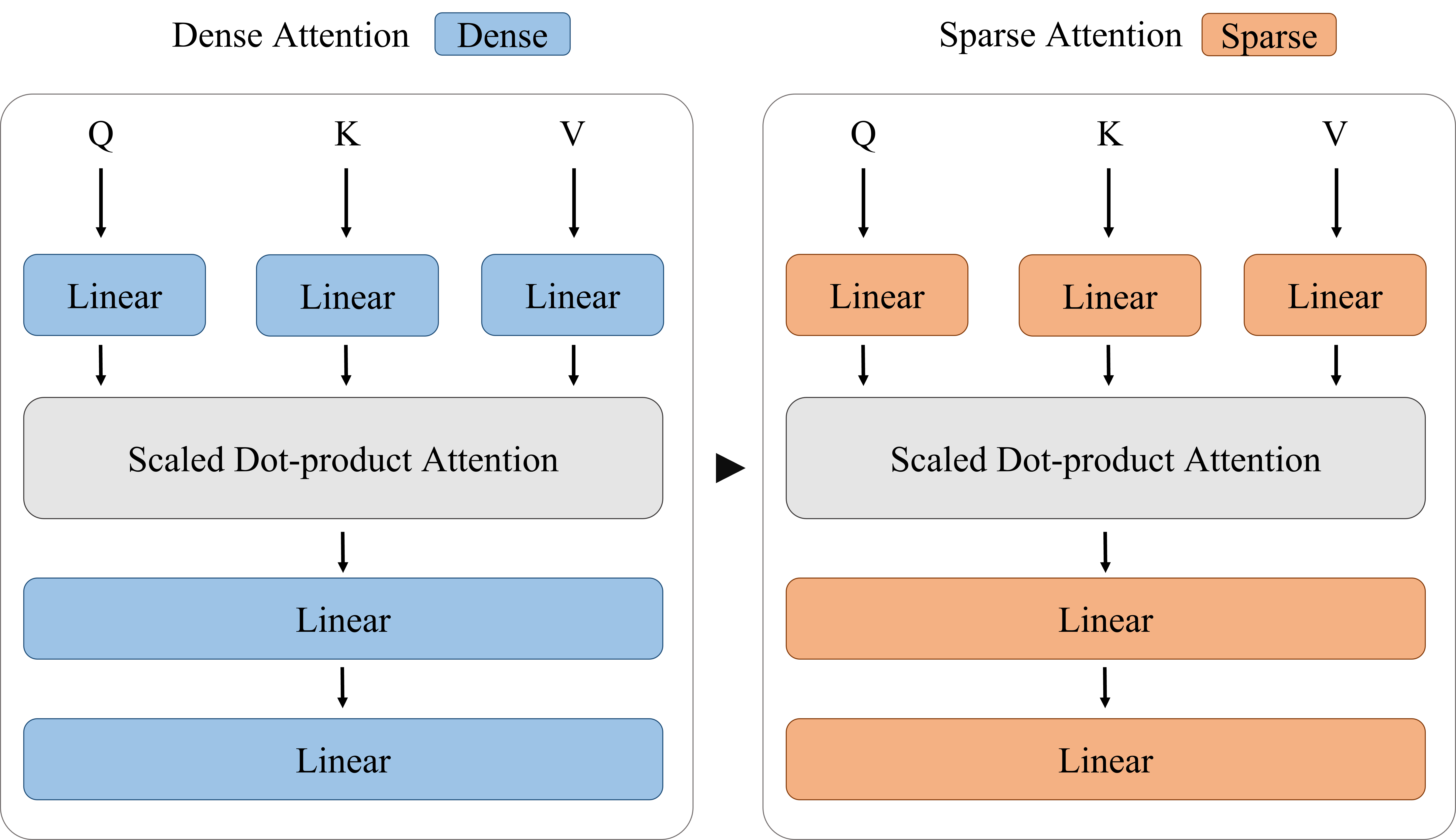}
        \end{minipage}
    }
    
    \subfigure[]{
        \begin{minipage}[t]{\linewidth}
            \centering
            \includegraphics[scale=0.22]{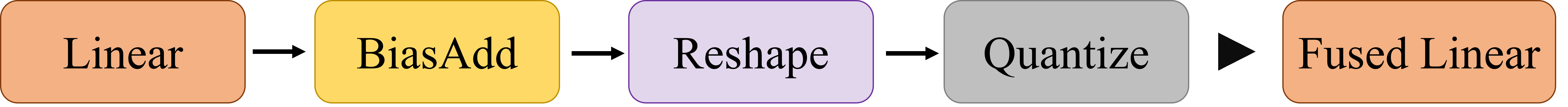}
        \end{minipage}
    }

    \subfigure[]{
        \begin{minipage}[t]{\linewidth}
            \centering
            \includegraphics[scale=0.22]{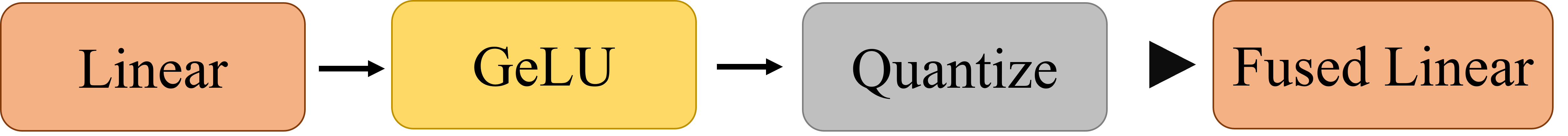}
        \end{minipage}
    }

    \subfigure[]{
        \begin{minipage}[t]{\linewidth}
            \centering
            \includegraphics[scale=0.22]{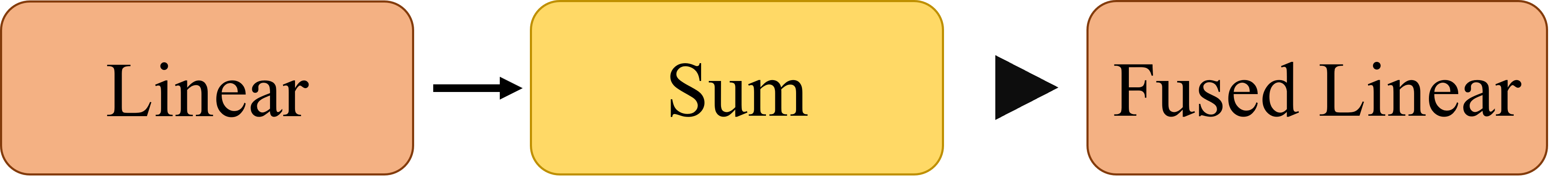}
        \end{minipage}
    }
    \centering
    \caption{Sparse attention and sparse Linear fusions. (a) Sparse attention vs. dense attention. All the Linear operators are converted from dense to sparse. Scaled dot-product attention is dense as there is no computation operators with the fixed weights. (b), (c), and (d) shows the sparse Linear fusion with the post-operators such as BiasAdd, Reshape, GeLU, Sum, Quantize etc. Specially, Quantize operator is introduced by INT8 quantization.}
    \label{fig:sparse_attention}
\end{figure}

\begin{multicols}{2}

Note that for the weight with 4 non-dividable sparsity dimension or 4 non-dividable non-zero blocks, the additional padding is needed to satisfy the accumulation dimensions for VNNI. For simplicity, we omit the special handling of padding in the sparse GEMM kernel implementation.

To scale the sparse GEMM kernel from single thread to multi-threads, we implement two-dimensions blocking strategy along M and N dimensions. Blocking on M dimension allows different kernels to compute with the corresponding sliced weight. However, blocking on N dimension introduces some redundant inter-core communication cost. To avoid such overhead, we re-layout \textit{K x N} matrix into 3-dimensional matrix \textit{NUM\_BN x K x BN}, where \textit{BN} means block size along N dimensions, \textit{NUM\_BN} means number of micro blocks in N dimension. Then we apply the thread parallelism along the first dimension \textit{NUM\_BN} to balance the task for each thread.

\subsection{Sparse Transformer Attention}
\label{sec:sparse_attention}
Transformer attention is a common block for Transformer-based LMs. With the sparse GEMM kernel, we can easily replace dense Linear operator with sparse Linear operator as shown in Figure~\ref{fig:sparse_attention}(a) where the diagram is a simplified version of Transformer attention~\cite{vaswani2017attention} for better illustration of sparse attention. A more complete Transformer attention actually consists of other operators such as BiasAdd, Reshape, GeLU, Sum etc. Unfortunately, these operators usually lead to the overhead of memory traffic and slow down the model performance.

Operator fusion is useful in deep learning for reducing the redundant memory and kernel launch overhead. There are two widely-used approaches: 1) computation graph-based operator fusion~\cite{jia2019optimizing} and graph compiler-based solution~\cite{rotem2018glow}. We apply the computation graph-based operator fusion given two considerations: 1) graph-based fusion solution is more mature and well adopted in industry; 2) operator fusion patterns are very common in Transformer attention and there is no need to complicate the fusion by introducing graph compiler-based solution. To support operator fusion, we then categorize three main kinds of operators to be fused with Linear operator: element-wise operator, binary operator, and shape manipulation operator. Typically, element-wise and binary operators can be fused into previous operator by reusing the data stored in SIMD registers to avoid memory movement between register and memory. Shape manipulation operators (e.g., Reshape) only modify the properties of a tensor without any computation logic, and therefore can be easily fused with other operators.

We implement an algorithm to fuse post-operators with sparse Linear operator. For each computation operator (e.g., Linear) in the computation graph, we take it as the starting operator and apply depth first search to identify the operators to be fused based on pre-defined categories. The identified operators are added into post-operator chain. 

Moreover, we develop a lookup-table (LUT) based approach to accelerate element-wise operators in low precision. Basically, LUT is a typical key-value table. Taking unsigned 8-bit integer as an example, the key range is from 0 - 255; the value is also INT8 quantized from FP32 value through pre-calculation for the post-operator chain offline. With the prepared LUT, we can directly get an output of the post-operator chain through a simple lookup operation given an input, without actual calculation during model inference. The idea of LUT can also be applied to an individual operator where there are intensive element-wise operations. Algorithm~\ref{alg:lut} gives the details on LUT generation.

\begin{algorithm}[H]
    \caption{LUT generation}
    \label{alg:lut}
    \KwIn{bit\_width\ $bit\_width$, op\_chain $op\_chain$}
    \KwOut{$LUT$}
        $LUT=init(bit\_width)$

        $index\gets min\_index(bit\_width)$

        \While{$index\leq max\_index(bit\_width)$}{
        \For{$op\ in\ op\_chain$}{
        $x\gets op(x)$\;
        }
        $LUT(index) \gets x$

        $index = get\_next\_index()$
        }
        \KwRet $LUT$\;
    \end{algorithm}
    
\subsection{End-to-end Sparse Inference Framework}
\label{sec:e2e_sparse_accelerator}
We develop an end-to-end sparse deep learning inference framework to demonstrate the performance for Transformer-based LMs. Our framework has the similar architecture as other inference solutions which consist of three levels of optimizations: operator, graph, and runtime. 

Operator optimization requires the optimal kernels for sparse and dense operators. We describe sparse GEMM kernels in Section~\ref{sec:sparse_gemm}. For the remaining dense operators such as BatchMatmul, LayerNorm, Softmax, we also develop JIT-based kernels with tunable configurations (e.g., tiling size for better register utilization). We enable the cache mechanism to allow the first-jitted kernel to be reused during inference if the operator shape is unchanged.

Graph optimization includes three stages: 1) pre-optimization (e.g., constant folding, common sub-expression elimination), 2) quantization (e.g., 16-bit or 8-bit), and 3) back-end optimization. We focus more on quantization and back-end optimization, since most of Transformer-based LMs are well-designed and there is few opportunities in pre-optimization stage. On low precision optimization, we leverage Intel\textsuperscript{\tiny\textregistered} Neural Compressor\footnote{https://github.com/intel/neural-compressor} to generate INT8 models by using built-in accuracy-aware tuning capability. Back-end optimizations have been mostly described in Section~\ref{sec:sparse_attention}.

Runtime optimization requires an efficient memory allocator and thread scheduler. The default memory allocator usually creates a new buffer each time when receiving a memory allocation request, and therefore the data is less likely to be reused. To reduce such memory allocation overhead, we develop a custom memory allocator to maximize the buffer reuse and make the data more cache friendly. To further shrink the memory usage, we implement weight sharing that allows a single copy of weight to be shared across multiple instances running in parallel during inference. Moreover, we create a thread management to schedule the thread usage more effectively to support inter- and intra-instance execution. 

\section{Experimental Setup}
\label{sec:setup}
We describe the experimental setup including hardware settings, sparse models and hyper-parameters, and kernel-level and model-level benchmark configurations.

\subsection{Hardware Settings}
\label{sec:hw_settings}
We select two popular x86 CPU instances (24 cores) on AWS: c6i.12xlarge for Intel\textsuperscript{\tiny\textregistered} Xeon\textsuperscript{\tiny\textregistered} Ice Lake and c6a.12xlarge for AMD\textsuperscript{\tiny\textregistered} Eypc for performance measurement, since these two CPU types are well validated in popular GEMM libraries and deep learning inference solution. Turbo is enabled by default for both instances on AWS.
\end{multicols}

\onecolumn
\begin{table}[htbp]
\caption{Sparse models, sparsity ratio, approach, and accuracy (Acc). Delta is the difference between sparse and dense accuracy. Typically, delta $>=$ -1\% is required, and higher is better. DistilBERT (Squad v1.1) shows better delta due to distillation used in pre-trained stage}
\label{tab:models}
\centering
\begin{tabular}{llclccc}
               \toprule
               Model & Dataset & Sparsity Ratio & Approach & Acc (Dense) & Acc (Sparse) & Acc (Delta) \\
               \midrule
               BERT-Mini & Squad v1.1 & 80\% & Dense + fine-tuned & 76.87\% & 76.27\% & -0.78\% \\
               BERT-Mini & MRPC	& 90\%	& Dense + fine-tuned & 87.52\% & 87.21\% & -0.36\% \\
               BERT-Mini & SST-2 & 90\%	& Dense + fine-tuned & 87.61\% & 86.92\% & -0.79\% \\
               DistilBert & Squad v1.1 & 80\% & Sparse + pre-trained & 85.8\% & 86.8\% & 1.17\% \\
               DistilBert & MRPC & 90\%	& Dense + fine-tuned & 88.85\% & 88.65\% & -0.23\% \\
               BERT-Base & Squad v1.1 & 80\%	& Sparse + pre-trained & 88.59\% & 88.67\% & 0.09\% \\
               BERT-Base & Squad v1.1 & 85\% & Sparse + pre-trained & 88.59\% & 88.03\% & -0.63\% \\
               BERT-Base & MRPC	& 80\% & Sparse + pre-trained & 90.5\% & 90.43\% & -0.08\% \\
               BERT-Base & MRPC	& 85\% & Sparse + pre-trained & 90.5\% & 89.63\% & -0.96\% \\
               \bottomrule
\end{tabular}
\end{table}

\begin{multicols}{2}

\subsection{Sparse Models}
\label{sec:sparse_models}
We use two training approaches to generate the sparse models: 1) initialize the weight from a dense model and prune the model during fine-tuning for a downstream task (dense + fine-tuned as short), and 2) initialize the weight from a pre-trained sparse model, lock the sparsity pattern, and fine-tune the model for a downstream task (sparse + pre-trained as short). Table~\ref{tb:hp_distilbert} shows the hyper-parameters for DistilBERT (the others in Appendix~\ref{sec:hp_more}). All the sparse models use 4x1 sparsity pattern, which demonstrates that this sparsity pattern allows us to achieve high sparsity ratio while maintaining less than 1\% accuracy loss for our tested models. 

\begin{table}[H]
 \caption{Hyper-parameters for sparse DistilBERT}
 \label{tb:hp_distilbert}
\centering
\begin{tabular}{lc}
               \toprule
               Hyper-parameter & DistilBERT (Squad 80\%)\\
               \midrule
               Learning rate & 1.8e-4 \\
               Batch Size & 12 \\
               Weight decay & 0.01 \\
               Epochs & 8 \\
               Learning rate decay & Linear \\
               Warmup ratio & 0.05 \\
               Sequence length & 384 \\
               $\lambda_{MLM}$ & 0 \\
               $\lambda_{kd}$ & 1 \\
               Temperature & 2 \\
               \bottomrule
 \end{tabular}
 \end{table}
 
\subsection{Benchmark Configurations}
\label{sec:benchmark_configs}
We benchmark our framework against commonly used solutions shown in Table \ref{tb:sw_version}. We show both kernel-level and end-to-end performance results.

On sparse kernel benchmark, we use single thread and four threads to measure the performance on a set of GEMM shapes (totally 90) that are widely used in typical Transformer models. For oneMKL and TVM, we refer to the document and sample codes from the sparse libraries to develop our benchmark code. 

\begin{table}[H]
 \caption{Software version used for kernel or model benchmark}
 \label{tb:sw_version}
\centering
\begin{tabular}{lcc}
               \toprule
               Software & Version & Type \\
               \midrule
               oneMKL & 2022.1.0 & Kernel \\
               LIBXSMM & 1.17 & Kernel \\
               TVM & 0.9.0 & Kernel \\
               \midrule
               Neural Magic (Deep Sparse) & 1.1.0 & Model \\
               ONNX Runtime & 1.11.1 & Model \\
               PyTorch & 1.11 & Model \\
               \bottomrule
 \end{tabular}
 \end{table}
 
 On model benchmark, the goal is to achieve the maximum throughput under proxy latency constraint for production per each model. You can see from Table~\ref{tb:latency_bs} that the proxy latency constraint per model almost aligns with the number of encoder layers and the weight shape. Note that the measured best latency may exceed the latency constraint under certain configurations, so we show the throughput based on the best latency using all the possible cores.
 
 \begin{table}[H]
 \caption{Model name, proxy latency constraint for production, number of encoder layers, and weight shape}
 \label{tb:latency_bs}
\centering
\begin{tabular}{lccl}
               \toprule
               Model & Latency & Encoder Layers & \makecell{Weight \\ Shape} \\
               \midrule
               BERT-Mini & 1 ms & 4 & \makecell{256x256 \\ 256x1024 \\ 1024x256} \\
               \midrule
               DistilBERT & 10 ms & 6 & \makecell{768x768 \\ 768x3072 \\ 3072x768}  \\
               \midrule
               BERT-Base & 20 ms & 12 & \makecell{768x768 \\ 768x3072 \\ 3072x768} \\
               \midrule
               BERT-Large & 50 ms & 24 & \makecell{1024x1024 \\ 1024x4096 \\ 4096x1024} \\
               \bottomrule
 \end{tabular}
 \end{table}
\end{multicols}

\onecolumn

\begin{figure}[htbp]
    \centering
        
    \subfigure[]{
        \begin{minipage}[t]{\linewidth}
            \centering
            \includegraphics[width=6in, height=2.95in]{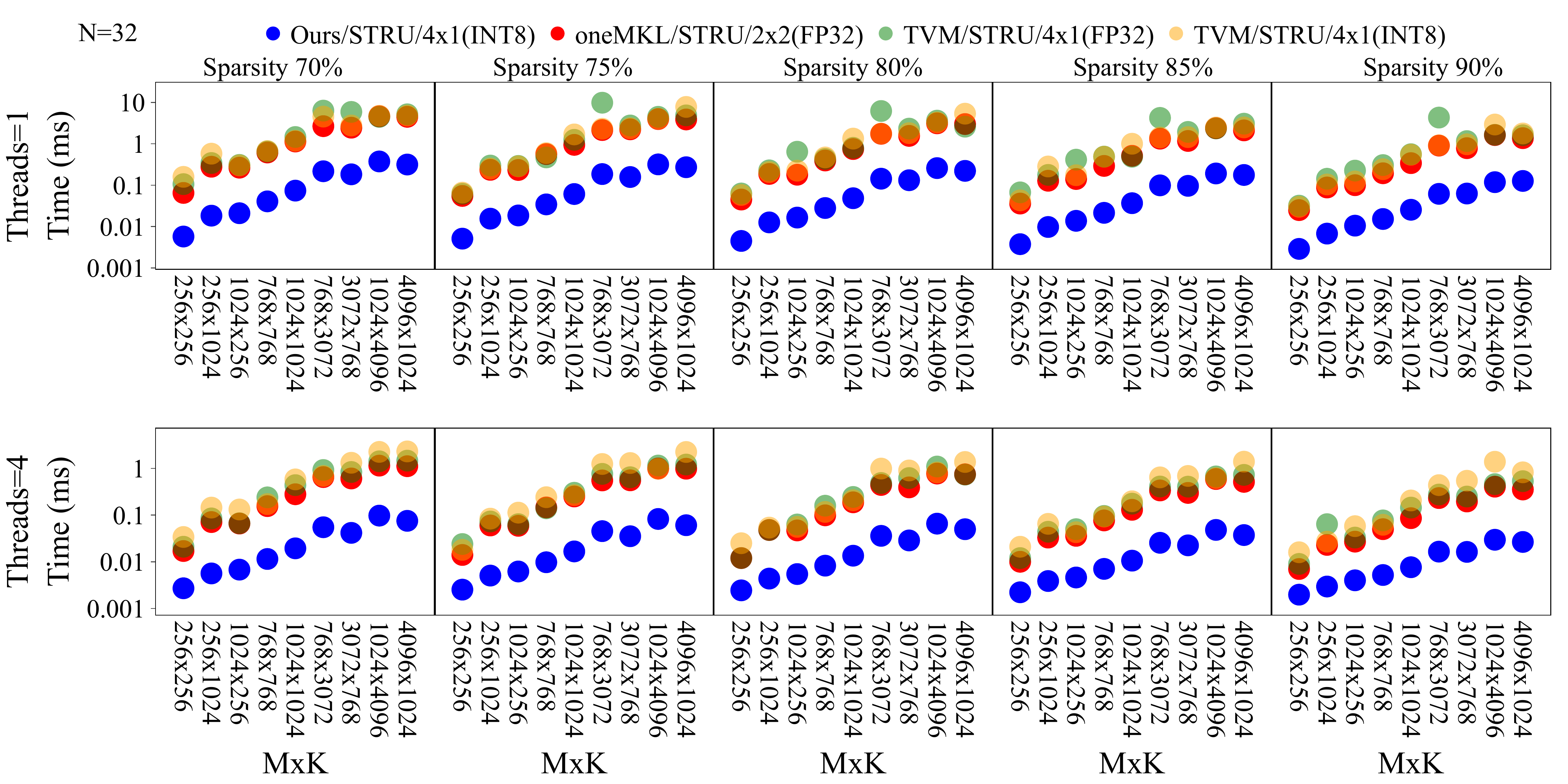}
        \end{minipage}
    }
    \vskip -0.05in
    \subfigure[]{
        \begin{minipage}[t]{\linewidth}
            \centering
            \includegraphics[width=6in,height=2.95in]{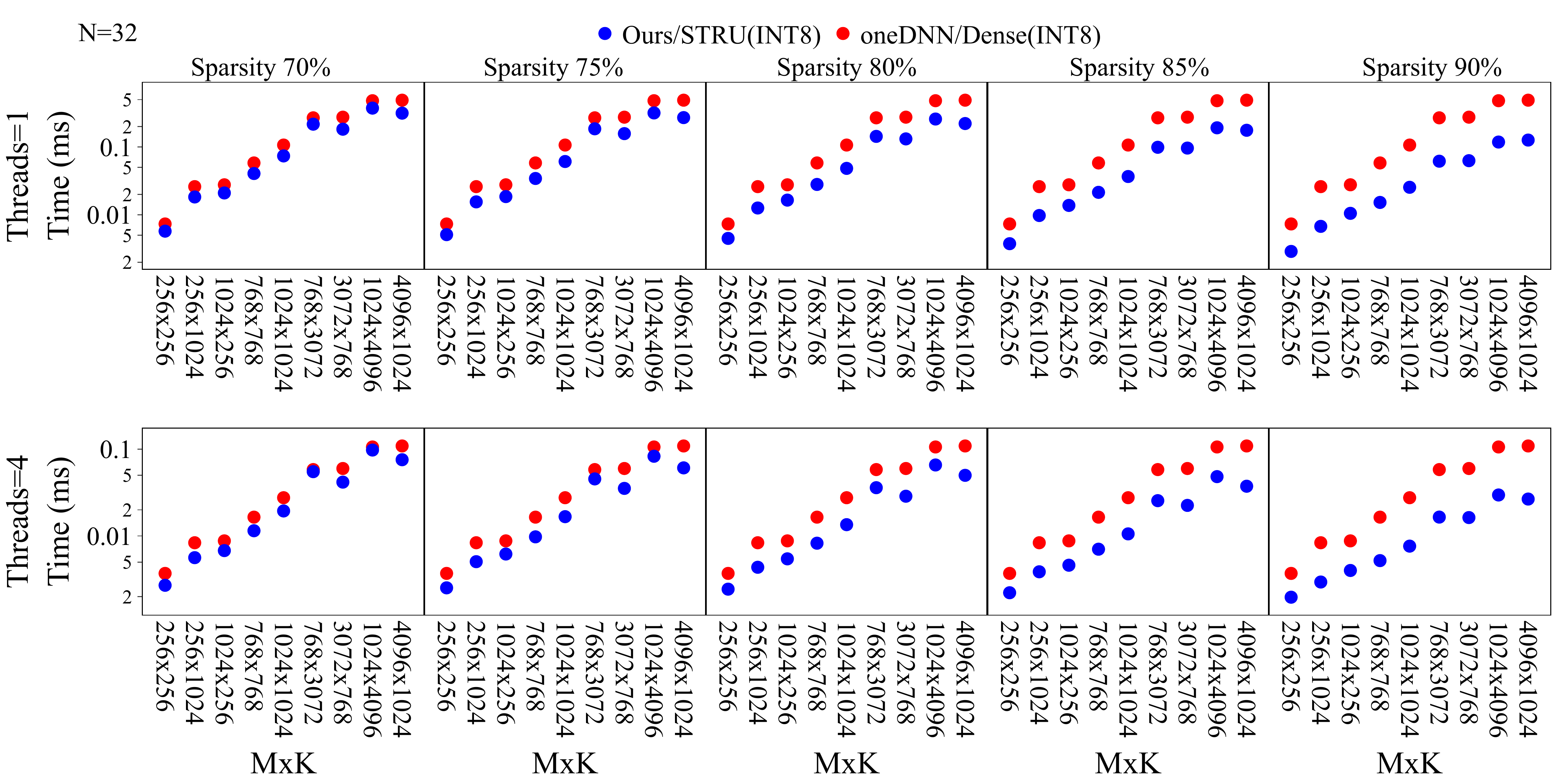}
        \end{minipage}
    }
    \centering
    \caption{Sparse GEMM kernel performance (N = 32). STRU and UNSTRU is structured and unstructured sparsity. (a) Comparing sparse GEMM on same block size across libraries, sparsity patterns (4x1, 2x2), sparsity ratios (70\% - 90\%), data types, and threads; (b) Comparing sparse with dense GEMM. X-axis is GEMM shape MxK, and Y-axis is $log_{10}$ based execution time on single or four threads}
    \label{fig:sparse_gemm_kernel}
\end{figure}

\begin{table}[htbp]
\caption{Geomean and maximum (Max) performance for our sparse GEMM kernels}
\label{tb:sparse_gemm_kernel_perf_n32}
\centering
\begin{tabular}{llllcc}
               \toprule
               Sparse Library & Sparsity Pattern & Sparsity Ratio & Data Type & \makecell{Thread 1 \\Geomean / Max} & \makecell{Thread 4 \\Geomean / Max} \\
               \midrule
               Ours vs. oneMKL & STRU 4x1 vs. STRU 2x2 & 70\% - 90\% & INT8 vs. FP32 & 12.7x / 16.4x & 10.9x / 16.2x \\
               Ours vs. TVM & STRU 4x1 vs. STRU 4x1 & 70\% - 90\% & INT8 vs. FP32 & 19.2x / 65.6x & 14.1x / 22.6x \\
               Ours vs. TVM & STRU 4x1 vs. STRU 4x1 & 70\% - 90\% & INT8 vs. INT8 & 16.5x / 31.5x & 18.7x / 47.0x \\
               \midrule
               Ours vs. oneDNN & STRU 4x1 vs. Dense & 70\% for Ours & INT8 vs. INT8 & 1.4x /
               1.6x & 1.3x / 1.5x \\
               Ours vs. oneDNN & STRU 4x1 vs. Dense & 75\% for Ours & INT8 vs. INT8 & 1.6x / 1.8x & 1.5x / 1.8x \\
               Ours vs. oneDNN & STRU 4x1 vs. Dense & 80\% for Ours & INT8 vs. INT8 & 1.9x / 2.2x & 1.8x / 2.2x \\
               Ours vs. oneDNN & STRU 4x1 vs. Dense & 85\% for Ours & INT8 vs. INT8 & 2.5x / 2.9x & 2.3x / 2.9x \\
               Ours vs. oneDNN & STRU 4x1 vs. Dense & 90\% for Ours & INT8 vs. INT8 & 3.6x / 4.4x & 3.1x / 4.1x \\
               \bottomrule
\end{tabular}
\end{table}

\begin{table}[htbp]
\caption{Geomean and maximum (Max) performance on sparse BERT-Mini (90\% sparsity ratio), DistilBERT (80\%), BERT-Base (80\%), and BERT-Large (80\%) on all the sequence lengths (16 - 384). ONNX RT is short for ONNX Runtime}
\label{tab:sparse_model_perf}
\centering
\begin{tabular}{llcccc}
               \toprule
\multicolumn{1}{l}{Inference Solution}       & CPU vs. CPU & \makecell{BERT-Mini 90\% \\ Geomean / Max} & \makecell{DistilBERT 80\% \\ Geomean / Max} & \makecell{BERT-Base 80\% \\ Geomean / Max} & \makecell{BERT-Large 80\% \\ Geomean / Max} \\
               \midrule
\multirow{2}{*}{Ours vs. Neural Magic} & Xeon vs. Xeon & - & 1.4x/1.5x & 1.3x/1.5x & 1.4x/1.8x            \\
                                       & Xeon vs. Eypc & - & 3.7x/5.0x & 3.2x/4.0x & 3.9x/7.3x             \\
               \midrule 
\multirow{2}{*}{Ours vs. ONNX RT} & Xeon vs. Xeon & 16.8x/37.7x & 6.1x/10.7x & 7.3x/11.3x & 6.5x/10.0x \\
                                       & Xeon vs. Eypc & 76.3x/345.9x & 12.9x/17.9x & 14.9x/21.0x & 13.7x/20.4x \\
               \midrule
\multirow{2}{*}{Ours vs. PyTorch     } & Xeon vs. Xeon & 32.5x/72.7x & 16.9x/24.3x & 10.8x/13.4x & 8.1x/10.7x \\
                                       & Xeon vs. Eypc & 73.5x/309.1x & 22.5x/36.8x & 21.0x/29.1x & 18.6x/29.0x \\
                \bottomrule
\end{tabular}
\end{table}
 
\begin{multicols}{2}

\section{Results}
\label{sec:results}

\subsection{Sparse Kernel Performance}
\label{sec:sparse_kernel_perf}
We measure the sparse GEMM kernel performance on Xeon based on benchmark configurations described in Section~\ref{sec:benchmark_configs}. Table~\ref{tb:sparse_gemm_kernel_perf_n32} shows the results among the sparse libraries (ours, oneMKL, TVM) where N is 32 as an example. Our sparse GEMM kernel outperforms the other sparse libraries in all the benchmark configurations. There are three main results based on the sample GEMM shapes:

\begin{itemize}
  \item Ours vs. other sparse libraries (Structured sparsity with same pattern or same number of block elements e.g., 2x2): demonstrate 10x - 12x for geomean performance and 16x for maximum over oneMKL (Structured 2x2); 14x - 19x for geomean and 22x - 64x for maximum over TVM
  \item Our sparse GEMM kernel vs. dense GEMM kernel (of oneDNN) shows 1.4x to 3.6x for geomean performance and 1.6x to 4.4x for minimum performance on single thread, and the similar performance on four threads
  \item Our sparse library shows the performance close to linear linear 90\%+ scaling from single to four threads
\end{itemize}

 Due to the space limitation, a more comprehensive performance comparison is described in Appendix~\ref{sec:sparse_gemm_kernel} considering sparse (structured vs. unstructured) and dense, different sparse block sizes (4x1, 2x2, 4x4, 16x1), and 90 GEMM shapes (N from 16 to 384).

\subsection{Sparse Model Performance}
\label{sec:sparse_model_perf}
We describe how to generate a FP32 sparse model in Section~\ref{sec:setup}. To demonstrate our sparse GEMM kernel, we need to generate the INT8 sparse model. We leverage Intel\textsuperscript{\tiny\textregistered} Neural Compressor, which is an open-source model compression tool offering accuracy-aware tuning capability, and produce the INT8 sparse model that can meet the accuracy criteria (relative loss less than 1\%). The sparse models generated for our sparse accelerator are also used for ONNX Runtime and PyTorch. Note that how to quantize these sparse models is out of scope in this paper, but the quantization recipes and instructions will be published on Github along with the other source code.

Neural Magic has its own sparse model zoo\footnote{https://sparsezoo.neuralmagic.com/} which provides the quantized model using unstructured or structured pruning on typical neural networks, including structured sparse model with block pattern 4x1 for DistilBERT (sparsity ratio 80\%) and BERT-Base (sparsity ratio 80\%), and unstructured sparse model for BERT-Large (sparsity ratio 80\% and 90\%), which are used for our benchmark. Note that for BERT-Mini, we skip it in the performance comparison as there is no published model in the sparse model zoo; for BERT-Large, we generate a structured sparse model with pattern 4x1 for our sparse accelerator based on the same sparsity ratio of their unstructured one.

\begin{figure}[htbp]
\centering
    \includegraphics[width=6in]{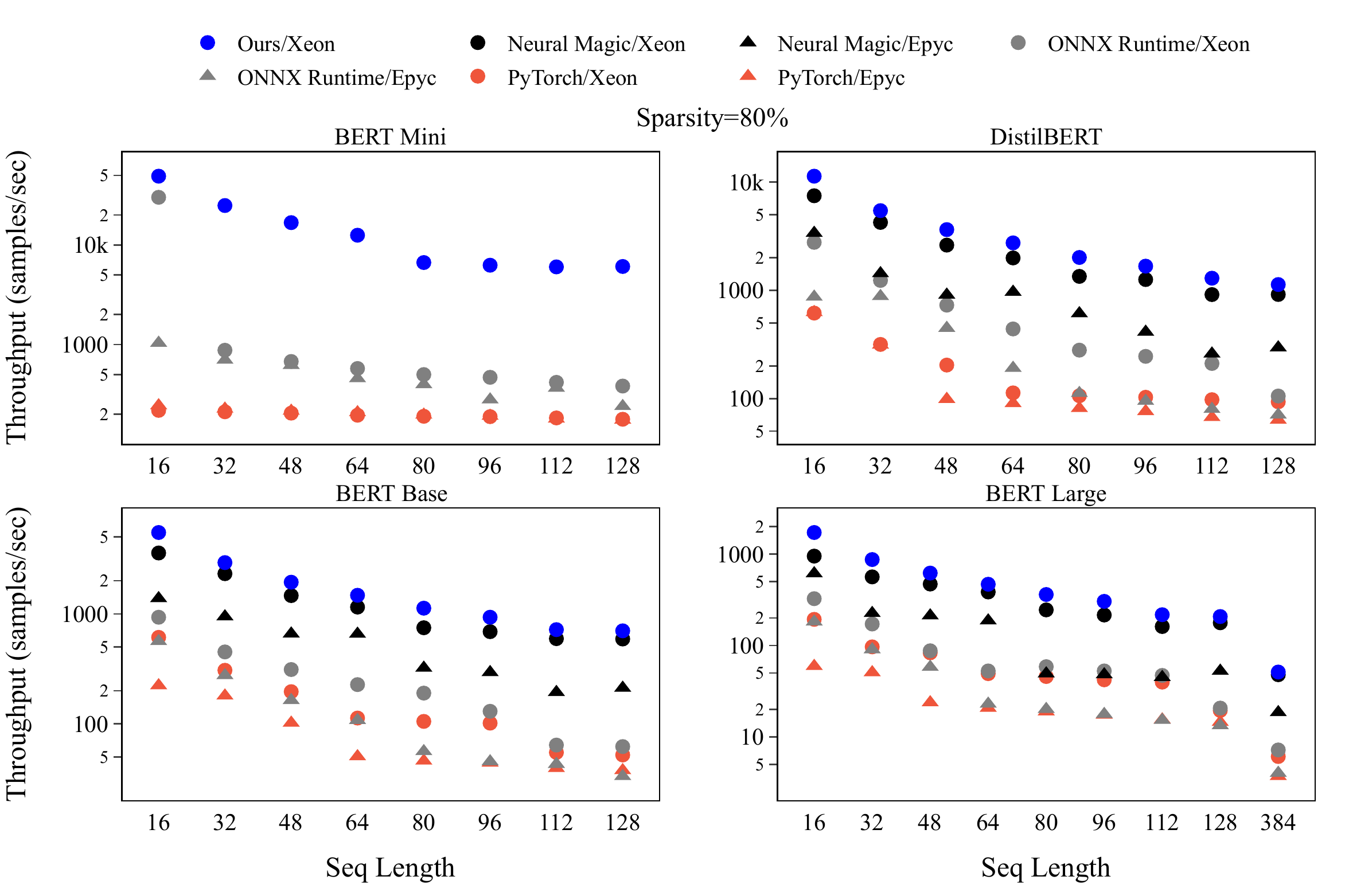}
    \caption{Sparse model performance (maximum throughput under latency constraint (in Table~\ref{tb:latency_bs})}
    \label{fig:sparse_model_performance}
\end{figure}

To the best of our knowledge, our sparse accelerator is the first one to demonstrate the performance on typical Transformer-based models across various downstream tasks. The results are presented in Table~\ref{fig:sparse_model_performance}. Our solution outperforms Neural Magic by 1.3x - 1.4x (geomean) and 1.5x - 1.8x (maximum), ONNX Runtime by 6x - 16x (geomean) and 10x - 37x (maximum), and PyTorch by 8x - 32x (geomean) and 10x - 72x (maximum) on same Xeon instance across different models. Figure \ref{fig:sparse_model_performance} shows the performance of each sparse model per difference sequence length. More interestingly, we also report the performance on Eypc instance which is also being used for inference.
  
\end{multicols}

\onecolumn

\begin{multicols}{2}

\subsection{Discussions}
\label{sec:perf_discussions}
We present the performance of sparse GEMM kernel and model performance in Section~\ref{sec:sparse_kernel_perf} and~\ref{sec:sparse_model_perf} and provide additional observations as below. 

On sparse GEMM libraries, oneMKL provides the best OOB experience to run the sparse kernel performance without additional tuning; TVM might be sub-optimal for AVX512 and therefore it shows the overall worse performance than the other two libraries; LIBXSMM provides sample code for sparse kernel benchmark while it does not support structured sparse GEMM kernels. On dense GEMM library, oneDNN shows decent performance on a wide range of GEMM shapes and good scaling from single to four threads. 

On end-to end sparse inference, we also measure the maximum throughput without latency constraints and minimal latency besides the default proxy production measurement. Similar to maximum throughput under proxy latency constraint, our sparse accelerator outperforms the other inference solutions both in maximum throughput without latency constraint and minimal latency in nearly all the configurations (shown in Figure~\ref{fig:workload_throughput_latency} in Appendix). In addition, to understand the performance on sparse model more thoroughly, we generate the sparse models based on different sparsity ratio from 70\% to 90\% with 5\% stride. Figure~\ref{fig:sparse_model_perf_sp_ratio} in Appendix shows almost linear scaling in performance on DistilBERT, BERT-Base, and BERT-Large among all the sequence lengths, but some negative scaling on BERT-Mini due to unbalanced thread/task scheduling on some corner GEMM shapes.

\section{Summary and future work}
\label{sec:summary}
In this paper, we developed an end-to-end solution for Transformer-based LMs inference with structured sparsity and quantization. Our SpMM kernel outperforms the existing sparse libraries (oneMKL, TVM, and LIBXSMM) by an order of magnitude on a wide range of shapes under representative sparsity ratios (70\%, 75\%, 80\%, 85\%, 90\%). We demonstrate large speedups on typical Transformer-based models (Bert-Mini, DistilBERT, Bert-Base, and BERT-Large) on CPUs: up to 1.5x and 4.9x over Neural Magic on same Xeon instance and different instances (Xeon vs. Eypc), up to 37x - 72x over ONNX Runtime and 309x - 345x over PyTorch from same to different CPU instance.

As future work, we plan to extend our software support to other CPU architectures (e.g., ARM) and contribute our open-source solution to the Transformer ecosystem. Moreover, we plan to extend the benchmark to provide Transformer users with the deployment choices for production with respect to the performance per dollar on cloud.

\bibliography{main}
\bibliographystyle{mlsys2023}
\end{multicols}

\appendix

\onecolumn

\section{Hyper-parameters for more sparse model}
\label{sec:hp_more}

\begin{table}[htbp]
 \caption{Hyper-parameters for sparse BERT-Mini}
 \label{tb:hp_more}
\centering
\begin{tabular}{lccc}
               \toprule
               Hyper-parameter & \makecell{BERT-Mini \\(Squad 80\%)} & \makecell{BERT-Mini \\(MRPC 90\%)} & \makecell{BERT-Mini \\(SST-2 90\%)} \\
               \midrule
               Learning rate & 4.5e-4 & 1e-3 & 5e-5\\
               Batch Size & 16  & 16 & 16 \\
               Weight decay & 1e-7 & 1e-3 & 5e-5\\
               Epochs & 10 & 15 & 15 \\
               Learning rate decay & Linear & Constant & Constant\\
               Warmup ratio & 0.018 & 0 & 0 \\
               Sequence length & 384 & 128 & 128 \\
               $\lambda_{kd}$ & 4.5 & 2.0 & 2.0  \\
               Temperature & 2 & 2 & 2\\
               \bottomrule
 \end{tabular}

 \caption{Hyper-parameters for sparse BERT-Base}
 \label{tb:hp}
\centering
\begin{tabular}{lcccc}
               \toprule
               Hyper-parameter & \makecell{BERT-Base \\Squad 80\%)} & \makecell{BERT-Base\\ (Squad 85\%)} & \makecell{BERT-Base \\(MRPC 80\%)} & \makecell{BERT-Base \\(MRPC 85\%)} \\
               \midrule
               Learning rate & 1e-4 & 1.5e-4 & 1e-4 & 1.5e-4 \\
               Batch Size & 12 & 12 & 32 & 32 \\
               Weight decay & 0 & 0 & 0 & 0 \\
               Epochs & 8 & 8 & 3 & 3 \\
               Learning rate decay & Linear & Linear & Linear & Linear \\
               Warmup ratio & 0 & 0 & 0 & 0 \\
               Sequence length & 384 & 384 & 128 & 128 \\
               $\lambda_{kd}$ & 1.0 & 1.0 & 1.0 & 1.0 \\
               Temperature & 2 & 2 & 2 & 2 \\
               \bottomrule
 \end{tabular}
\end{table}

\section{Sparse GEMM kernel performance}
\label{sec:sparse_gemm_kernel}
\begin{figure}[htbp]
\centering
    \includegraphics[scale=0.36]{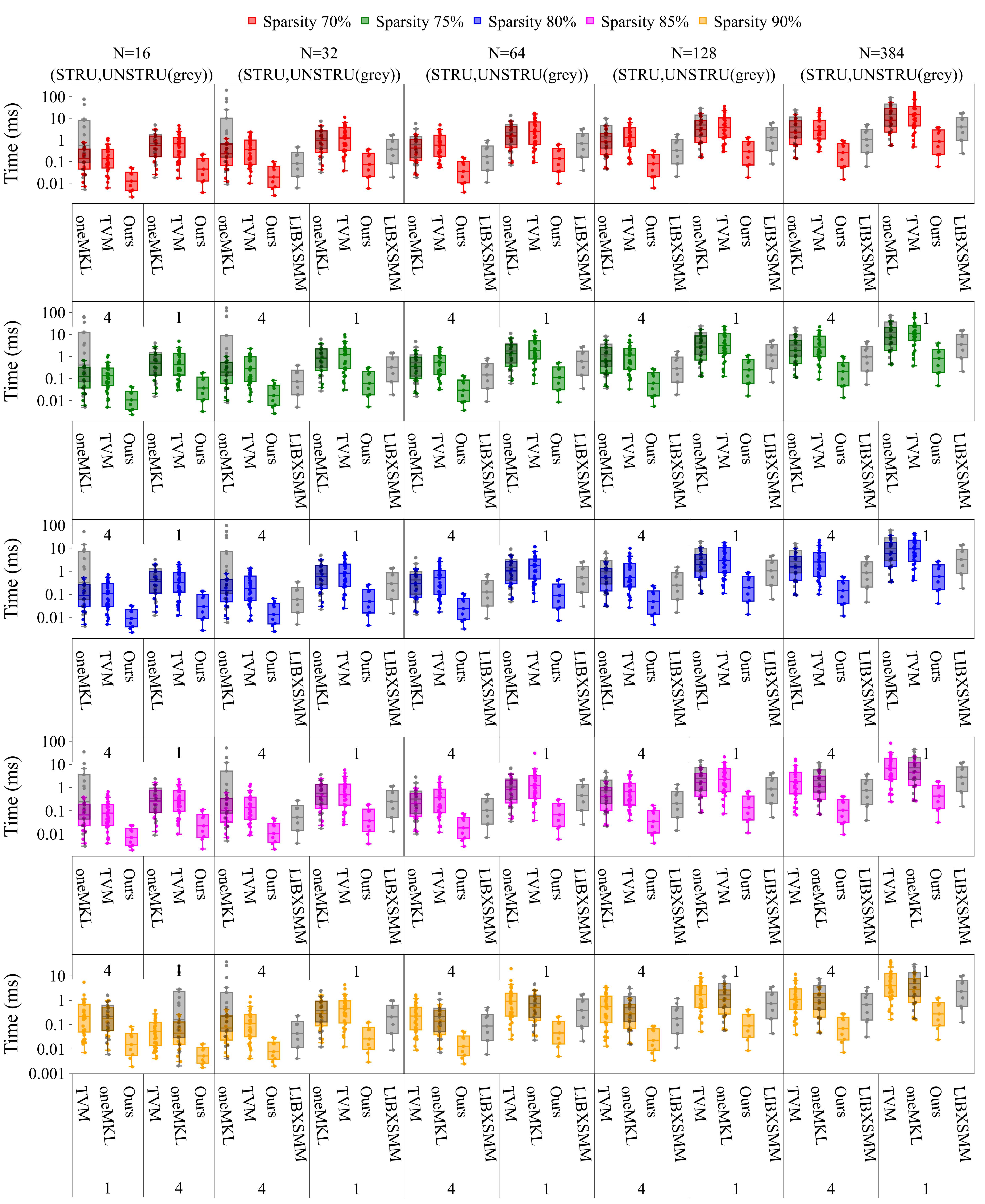}
    \caption{Sparse GEMM kernel performance: a high-level overview. Besides structured sparse GEMM kernel with same block size, we also compare: 1) structured SpMM with different or more aggressive block size: our kernel still shows competitive performance up to 12x over oneMKL (4x4) and 36x over TVM (16x1); and 2) unstructured SpMM: our kernel also shows up to 16x over oneMKL and 31x over LIBXSMM}
    \label{fig:sparse_gemm_perf_overview}
\end{figure}

\begin{table}[htbp]
\caption{Sparse GEMM kernel performance of all 90 shapes across libraries, sparsity patterns, data types, and threads}
\label{tb:sparse_gemm_kernel_perf}
\centering
\begin{tabular}{lllcc}
               \toprule
               Library & Sparsity & Data type & \makecell{Thread 1 \\Geomean / Maximum} & \makecell{Thread 4 \\Geomean / Maximum} \\
               \midrule
               Ours vs. oneMKL & Stru 4x1 vs. Unstru & INT8 vs. FP32 & 8.8x / 20.4x & 8.1x / 22.4x \\
               Ours vs. oneMKL & Stru 4x1 vs. Stru 2x2 & INT8 vs. FP32 & 11.2x / 17.5x & 10.4x / 20.3x \\
               Ours vs. oneMKL & Stru 4x1 vs. Stru 4x4 & INT8 vs. FP32 & 8.4x / 12.1x & 8.1x / 12.9x \\
               Ours vs. TVM & Stru 4x1 vs. Stru 4x1 & INT8 vs. FP32 & 19.2x / 75.6x & 15.5x / 41.7x \\
               Ours vs. TVM & Stru 4x1 vs. Stru 16x1 & INT8 vs. FP32 & 13.0x / 104.8x & 11.6x / 71.5x \\
               Ours vs. TVM & Stru 4x1 vs. Stru 4x1 & INT8 vs. INT8 & 19.2x / 41.4x & 20.9x / 62.8x \\
               Ours vs. TVM & Stru 4x1 vs. Stru 16x1 & INT8 vs. INT8 & 7.6x / 25.3x & 8.2x / 36.6x \\
               Ours vs. TVM & Stru 4x1 vs. Unstru & INT8 vs. FP32 & 7.7x / 25.3x & 8.4x / 31.1x \\
               Ours vs. oneDNN & Stru 4x1 vs. Dense & INT8 vs. INT8 & 2.1x / 4.8x & 2.0x / 5.1x \\
               \bottomrule
\end{tabular}
\end{table}

\newpage

\section{Workload performance}
\label{sec:workload_perf}

\begin{figure}[htbp]
\centering
    \includegraphics[width=6in]{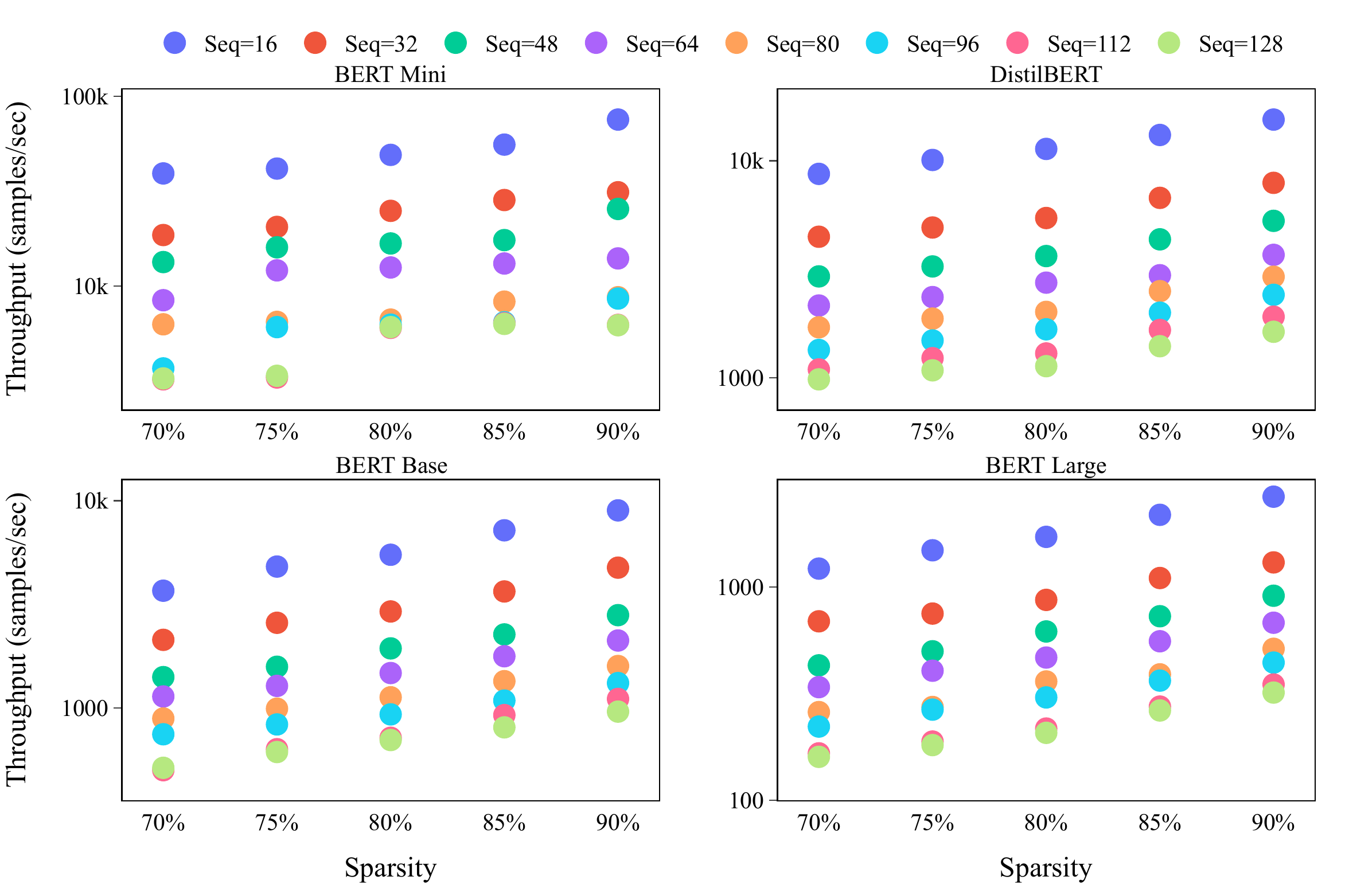}
    \caption{Sparse model performance scaling by sparsity ratio}
    \label{fig:sparse_model_perf_sp_ratio}
\end{figure}

\begin{figure}[htbp]
    \centering
    \subfigure[]{
        \begin{minipage}[t]{\linewidth}
            \centering
            \includegraphics[width=6in]{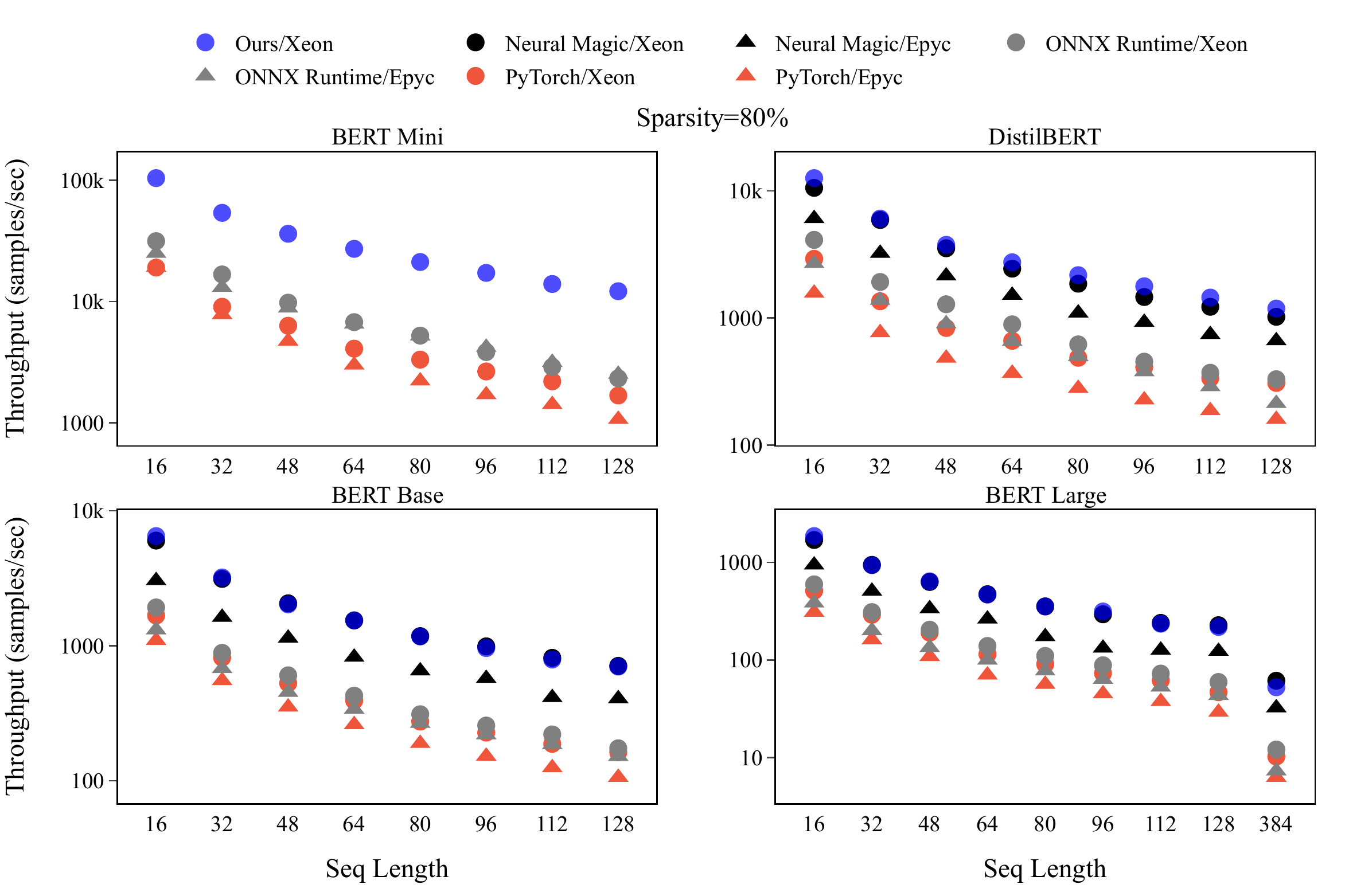}
        \end{minipage}
    }
    
    \subfigure[]{

        \begin{minipage}[t]{\linewidth}
            \centering
            \includegraphics[width=6in]{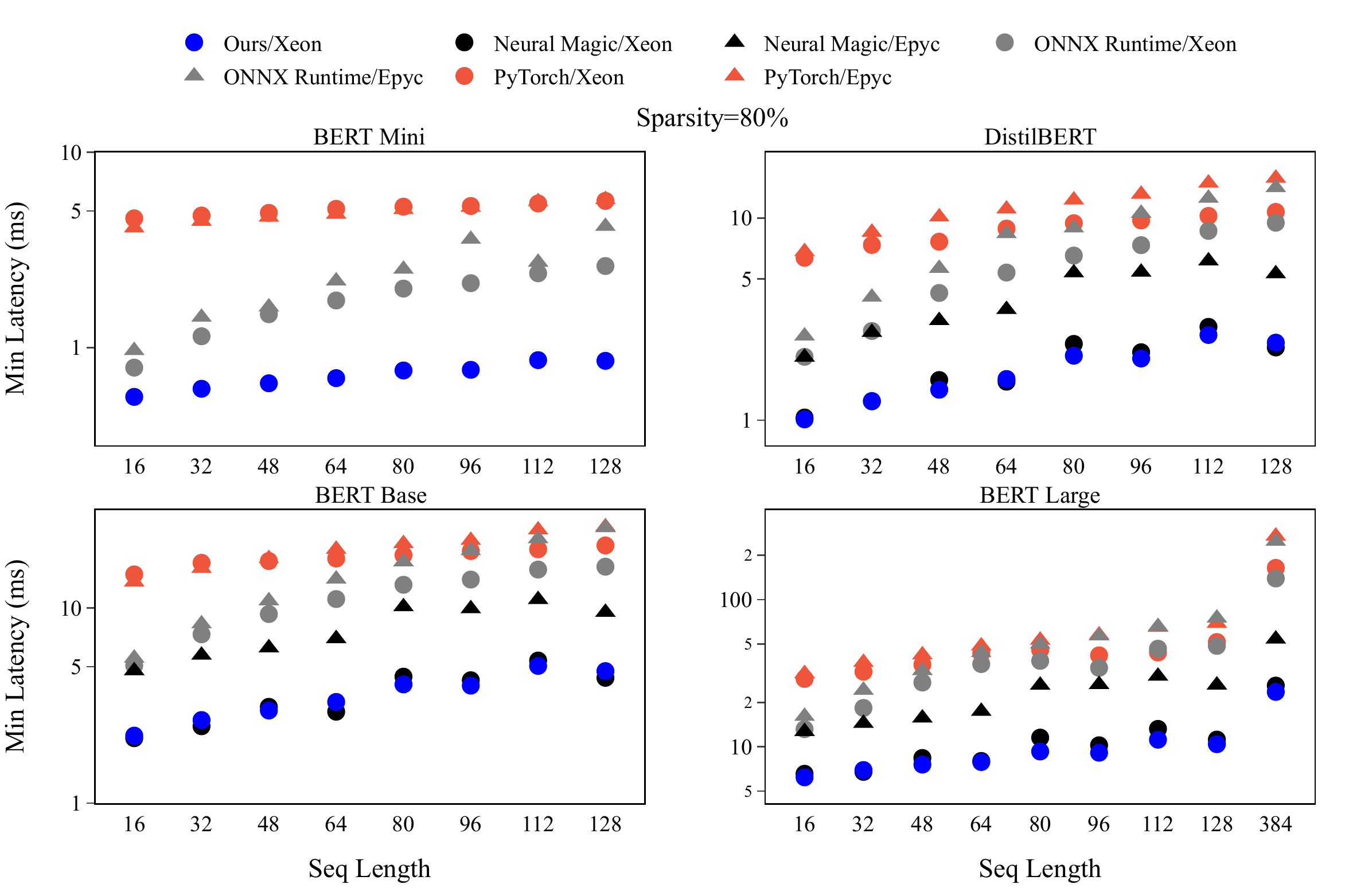}
        \end{minipage}
    }
    \centering
    \caption{Sparse model performance on maximum throughput without latency constraint (a) and minimal latency (b). X-axis is sequence length, and Y-axis is throughput with unit samples/sec (higher is better) in (a) and minimal latency with unit ms (lower is better) in (b), both of which are $log$ based number}
    \label{fig:workload_throughput_latency}
\end{figure}


\end{document}